\documentclass{article}
\usepackage{bm}
\usepackage{bbding}
\usepackage{spconf,amsmath,graphicx}
\usepackage{cite}
\usepackage{amsmath,amssymb,amsfonts}
\usepackage{algorithmic}
\usepackage{graphicx}
\usepackage{textcomp}
\usepackage{xcolor}
\usepackage{booktabs}
\usepackage{multirow}
\usepackage[normalem]{ulem}
\usepackage{booktabs}
\usepackage{wrapfig}
\usepackage{setspace}
\useunder{\uline}{\ul}{}
\usepackage{enumitem}
\setlist{nosep, leftmargin=14pt}
\usepackage[colorlinks,
            linkcolor=blue,
            anchorcolor=blue,
            citecolor=blue]{hyperref}

\usepackage{mwe} % to get dummy images

\title{Saliency-guided and Patch-based Mixup for Long-tailed Skin Cancer Image Classification}
\name{Tianyunxi Wei$^{1}$, Yijin Huang$^{1,2}$, Li Lin$^{1,3}$, Pujin Cheng$^{3}$, Sirui Li$^{1}$, Xiaoying Tang$^{1,4*}$}
\address{$^1$Department of Electronic and Electrical Engineering, Southern University of Science and Technology,\\ Shenzhen, China\\
$^2$School of Biomedical Engineering, University of British Columbia, \\Vancouver, Canada\\
$^3$Department of Electrical and Electronic Engineering, The University of Hong Kong,\\Hong Kong SAR, China\\
$^4$Jiaxing Research Institute, Southern University of Science and Technology,\\ Jiaxing, China}
\begin{document}
%\ninept
%
\maketitle
\begin{abstract}
Medical image datasets often exhibit long-tailed distributions due to the inherent challenges in medical data collection and annotation. In long-tailed contexts, some common disease categories account for most of the data, while only a few samples are available in the rare disease categories, resulting in poor performance of deep learning methods. To address this issue, previous approaches have employed class re-sampling or re-weighting techniques, which often encounter challenges such as overfitting to tail classes or difficulties in optimization during training. In this work, we propose a novel approach, namely \textbf{S}aliency-guided and \textbf{P}atch-based \textbf{Mix}up (SPMix) for long-tailed skin cancer image classification. Specifically, given a tail-class image and a head-class image, we generate a new tail-class image by mixing them under the guidance of saliency mapping, which allows for preserving and augmenting the discriminative features of the tail classes without any interference of the head-class features. Extensive experiments are conducted on the ISIC2018 dataset, demonstrating the superiority of SPMix over existing state-of-the-art methods. The source code is available at \url{https://github.com/Yancy10-1/SPMix}.
\end{abstract}
%
%\vspace{-0.3cm}
\section{Introduction}
\label{sec:intro}
%\vspace{-0.3cm}
% \begin{figure}[t]
% \centering
% \setlength{\abovecaptionskip}{-0.2cm}   %调整图片标题与图距离
% \setlength{\belowcaptionskip}{-1cm}   %调整图片标题与下文距离
% \includegraphics[width=\linewidth]{picture2.pdf}
% % %\vspace{-0.3cm}
% \caption{SPMix utilizes lesion-aware mixup ratio, which specifies the mixup ratio of each patch flexibly according to the average saliency score of each patch. It ensures to maximize the preservation of lesion regions within tail-class samples and replace the background of tail-classes with the rich background from head classes.} \medskip
% \label{mixupstrategy}
% % %\vspace{-0.7cm}
% \end{figure}

With the advent and development of deep learning, substantial advancements have been achieved in the domain of visual recognition, particularly in the realm of classification tasks\cite{sandler2018mobilenetv2,ma2018shufflenet}. However, supervised deep learning methods usually require balanced training datasets. Compared with natural image datasets, medical image data are more likely to exhibit long-tailed distributions, wherein head classes dominate most of the data, while tail classes have relatively limited samples. Such skewed datasets typically arise from the inherent challenges of collecting data for rare diseases, as well as the high cost of annotating medical images. Training deep learning models on such imbalanced datasets presents inherent challenges, as the low-frequency classes are susceptible to being overshadowed by the high-frequency classes, resulting in a notable deterioration in the performance of deep learning models.
\begin{figure}[t]
\centering
\setlength{\abovecaptionskip}{-0.2cm}   %调整图片标题与图距离
\setlength{\belowcaptionskip}{-1cm}   %调整图片标题与下文距离
\includegraphics[width=\linewidth]{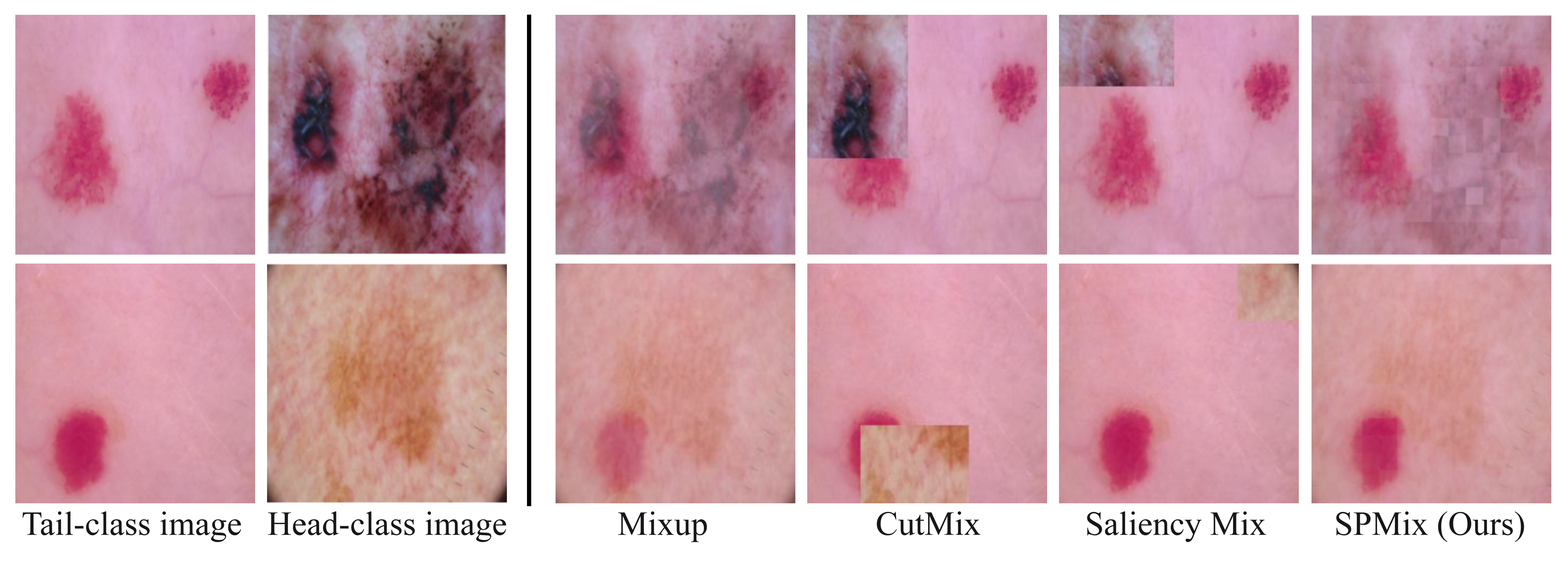}
\vspace{-0.7cm}
\caption{A comparison of different mixup methods. We visualize SPMix on the image level. As presented, SPMix ensures to preserve the discriminative features of tail-class samples, while other methods may compromise these features or exhibit inadequate levels of mixup.} \medskip
\label{compare}
\vspace{-0.7cm}
\end{figure}

\begin{figure*}[t]
\centering
\vspace{-1cm}
\includegraphics[width=\linewidth]{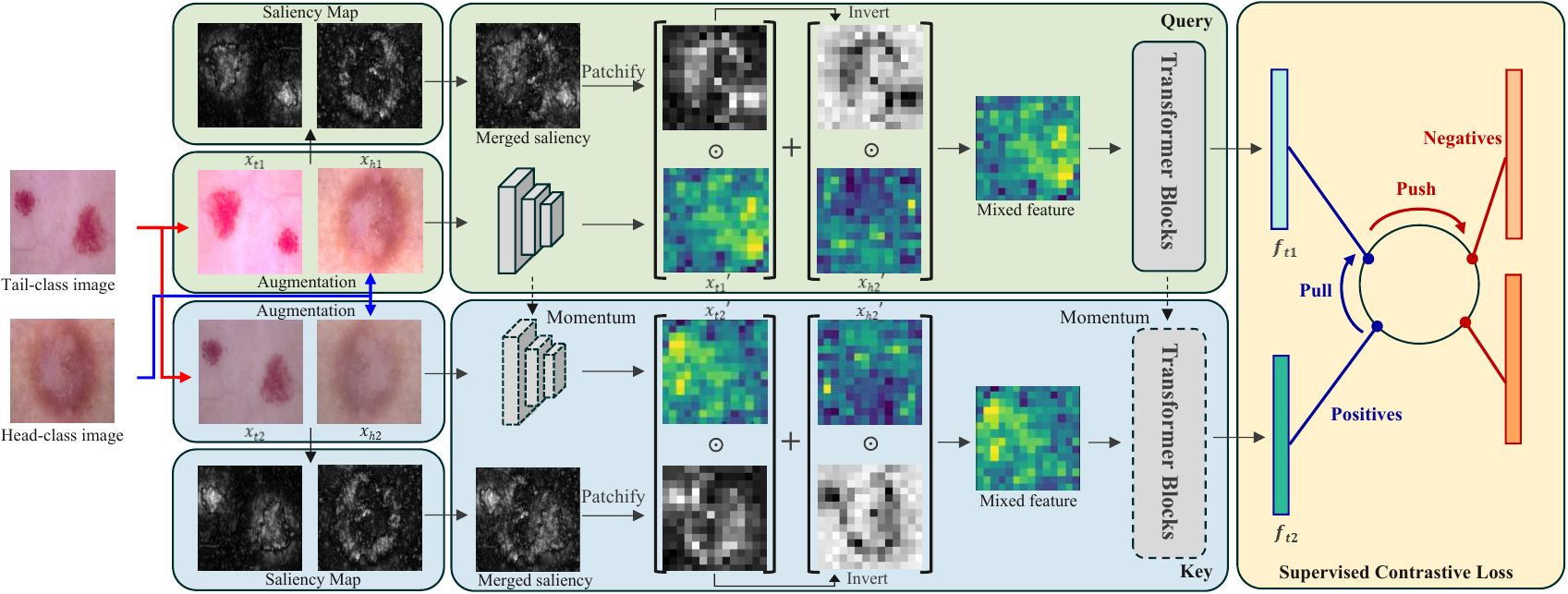}
\vspace{-0.6cm}
\caption{The overall SPMix framework. First, the tail-class image and head-class image are augmented differently to generate $(x_{t1}, x_{t2}), (x_{h1}, x_{h2})$. The pairs are respectively fed into a query and key encoder, which have the same architecture and the key encoder is driven by a momentum update with the query encode. The saliency map of the pairs are merged and patchified. The average value of each patch is used as the patch-based mixup ratio. The mixed features serve as the new positive pair and are utilized for the supervised contrastive loss.} 
\label{model}
\vspace{-0.4cm}
\end{figure*}

To overcome this challenge, previous work typically re-balanced the distribution of the original data through class re-balancing or re-weighting\cite{b4,b7,rebalancing1}, increasing the risk of overfitting in the tail classes or sacrificing the performance of the head classes. Recently a new line of work is proposed which uses a two-stage training process to decouple representation learning and classifier fine-tuning\cite{b9}: the first stage trains the feature extractor with the original data distribution, and then the classifier is fine-tuned with resampled data. Until supervised contrastive learning (SCL) \cite{b11} is introduced. PaCo\cite{cui2021parametric} introduces SCL equipped with a set of parametric class-wise learnable centers to address the long-tailed problem. BCL\cite{zhu2022balanced} alleviates these issues by introducing within-batch class average, class complement, and logit compensation to obtain a regular simplex configuration, but its performance is still limited by its small batch size. Data augmentation methods have shown promising performance in the field of computer vision. The mixup methods\cite{zhang2017mixup,yun2019cutmix,uddin2020saliencymix} are closely related to oversampling methods, and several recent long-tailed recognition methods have adopted mixup techniques to enrich the representation of tail classes. MiSLAS\cite{zhong2021improving} employed mixup during the Stage-1 training. However, applying vanilla mixup methods directly to medical data may lead to significant loss of important lesion information (Fig.~\ref{compare}).

In this paper, we propose an effective mixup strategy for long-tailed medical image classification, namely Saliency-guided and Patch-based Mixup (SPMix). Specifically, we introduce saliency map to guide patch-based feature-wise mixup, the saliency map plays a crucial role in clearly highlighting the salient region from the surrounding skin background. We emphasize the features of medical long-tailed images, particularly the rich context of the head classes. Given that the lesion regions often exhibit lumpiness and hold critical diagnostic value, we aim to generate new samples that better preserve these regions within the less-represented tail classes. To achieve this, we leverage the wealth of samples from the prominent head classes to compensate for the background components within the tail classes. We perform SPMix on the feature level to generate enriched samples of tail classes by merging the tail-class samples and head-class samples guided by the saliency map. Subsequently, a SCL based framework equipped with transformer blocks\cite{dosovitskiy2020image} is adopted for classification using the generated balanced dataset. In summary, our key contributions can be summarized as follows: (1) We propose a novel SCL framework with saliency-guided and patch-based mixup long-tailed classification task. We introduce saliency map into mixup, allowing us to generate tail-class samples that retain the diagnostic features of the tail classes. (2) Different from the vanilla mixup strategy, SPMix utilizes lesion-aware mixup ratio, which specifies the mixup ratio of each patch flexibly according to the saliency map. This strategy is more suitable for the characteristics of medical images. (3) Extensive experiments are conducted on ISIC2018 dataset, demonstrating the effectiveness of SPMix for long-tailed medical image classification.

\vspace{-0.1cm}
\section{METHODS}
\label{sec:format}
%\vspace{-0.3cm}
\subsection{Overall Framework}
The overall framework of our proposed SPMix is shown in Fig.~\ref{model}, which consists of three components: (1) By fusing the tail-class and head-class images, augmented images are generated to construct a balanced dataset. (2) Extract the features and mix the corresponding feature patches with saliency guidance. (3) Further encode the mixed features with transformer blocks and employ SCL for representation learning.

%\vspace{-0.3cm}
\subsection{Saliency Guidance}
In order to address the long-tailed problem in which excessive attention is paid to the head classes at the expense of the tail classes, we employ the head-class samples to compensate for the scarcity of tail-class samples. Our primary objective is to maximize the preservation of lesion regions within tail-class samples and replace the background of tail classes with the rich background from head classes. To this end, we introduce a saliency map to highlight the lesion regions in the images, with pixel values representing the degree of saliency\cite{huang2022ssit}. We utilize a static saliency detection method\cite{montabone2010human} that relies on the calculation of central-surround differences within images to determine saliency. This method is particularly well-suited for medical images with distinct intensity differences between lesion regions and backgrounds. As a result, it can be effectively applied to analyze skin cancer images\cite{wang2022superpixel}.

We derive the merged saliency map by considering the higher saliency scores at corresponding locations in both the tail-class samples and the head-class samples. This approach ensures that the salient regions specific to the head-class image are not incorporated in the augmented tail-class image, while concurrently retaining the diagnostics saliency of the tail-class image to the greatest extent. 
\vspace{-0.2cm}
\subsection{Lesion-aware Mixup Ratio}
Given that the saliency map remains constant, the fusion of the two images consistently yields the same result. Therefore, we introduce random noise superimposed on the previous merged saliency map. This augmentation serves a dual purpose: it elevates the complexity of network training while imparting a regularization effect, ultimately enhancing the model's generalization performance. Subsequently, we apply min-max normalization on the saliency map to ensure the saliency scores fall in the range of $[0,1]$. Some salient patches are more significant and their scores go up to 1, so that the corresponding mixup ratio is $1$ and then feature patches will not be mixed. To facilitate the blending process, we establish a threshold, represented as $\alpha$, and scores exceeding this threshold are clipped to it. This ensures that each feature patch is incorporated into the mixing process, and finally we derive the mixup ratio for each patch, denoted as $r$ corresponding to their respective positions,
\begin{equation}
r =\min(\alpha,\max (s_{h}, s_{t})),
% \vspace{-0.5cm}
\end{equation}
where $s_h$ and $s_t$ are  denoted as the saliency scores of head-class image and tail-class image.
\vspace{-0.2cm}
\subsection{Patch-based Mixup}
% 由于普通的mixup可能会扰乱病灶区域的重要信息，每张图片只有固定的一个融合比例，生成的新样本会引入头部类的病灶或者丢失部分尾部类病灶的信息，而绝大多数病灶区域都是块状的，所以我们将每个样本都分割成块状这样可以更可能保留尾部类样本病灶，每个块根据显著性来决定融合比例，the average saliency score of each patch is considered as the mixup ratio of that patch.越显著的块就是病灶部分，对应的融合比例就越低，而不显著的块就是背景区域，就用头部类样本的背景进行替换，这样灵活地融合更适应医学图像，
Due to the fact that vanilla mixup may disturb important information in the lesion regions and each sample only has a fixed mixup ratio, it results in the introduction of the lesion of head-class samples or the loss of some lesion information in the tail-class samples in the augmented samples. Most of the lesion regions are patch-shaped, so we utilize patch-based mixup with each sample to better preserve the lesion information of tail-class samples. The average saliency score of each patch is considered as the mixup ratio of that patch. The more significant patch is the lesion region, the higher the corresponding mixup ratio. The non-significant patch refers to the background region, which is substituted with the background from the head-class sample. This adaptable mixup strategy proves to be highly suitable for medical images. Our framework consists of two structurally identical networks, namely the query network and key network. The key network is driven by the momentum update of the query network during training. For each network, it typically contains an encoder CNN to generate feature maps for feature mixup and a sequence of transformer blocks to further encode the mixed features. Given one tail-class sample $x_t$ and one head-class sample $x_h$, data augmentation is first utilized to generate two augmented views for each sample, $(x_{t1}, x_{h1})$ and $(x_{t2}, x_{h2})$. Then, we compute the saliency-guided mixup ratio $r_1$ and $r_2$ for each pair. We generate new tail-class sample by 
\vspace{-0.2cm}
\begin{equation}
\begin{split}
{x_{t1}}^{'}= r_{1} * x_{t1} + (1 - r_{1}) * x_{h1}, \\
{x_{t2}}^{'} = r_{2}* x_{t2} + (1 - r_{2}) * x_{h2},
\end{split}
\vspace{-0.1cm}
\end{equation}
where $({x_{t1}}^{'}, {x_{t2}}^{'})$ are considered as a new positive pair of tail class for SCL. They are fed into the query and key transformer blocks respectively and we denote their outputs as $(f_{t1}, f_{t2})$. $(F_{t1}, F_{t2})$ is the new positive pairs of a batch unit.
\begin{figure}[t]
\begin{flushright}
\vspace{-0.1cm}
\centering
\setlength{\abovecaptionskip}{-0.2cm}   %调整图片标题与图距离
\setlength{\belowcaptionskip}{-1cm}   %调整图片标题与下文距离
\includegraphics[width=0.8\linewidth]{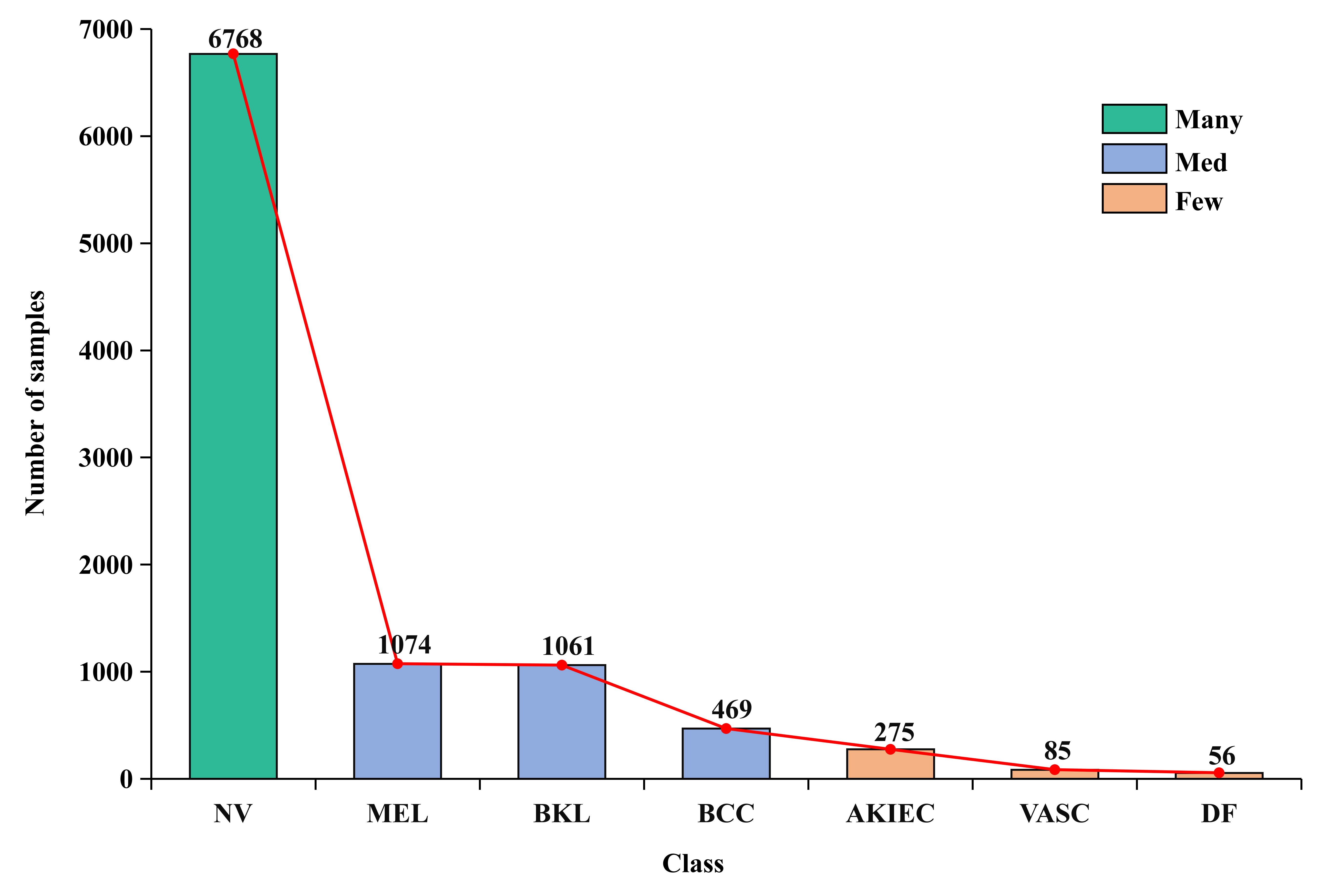}
\vspace{-0.3cm}
\caption{Long-tailed distributions of the ISIC2018.} \medskip
\label{dataset}
\vspace{-1.1cm}
\end{flushright}
\end{figure}

\subsection{Supervised Contrastive Loss}
With SPMix, we utilize the rich background of head-class samples to diversify the background of tail-class samples, thereby constructing a balanced dataset, which is then further trained using SCL. In SCL, positive samples are all images within the same class, and SCL can flexibly merge any number of positive values. It optimizes the consistency between these positive samples by comparing negative samples.
% \vspace{-0.1cm}
\vspace{-0.9cm}
\section{ EXPERIMENTS}
% \vspace{-0.3cm}
\subsection{Dataset and Evaluation}
\begin{figure}[b]
\vspace{-0.5cm}
\includegraphics[width=\linewidth]{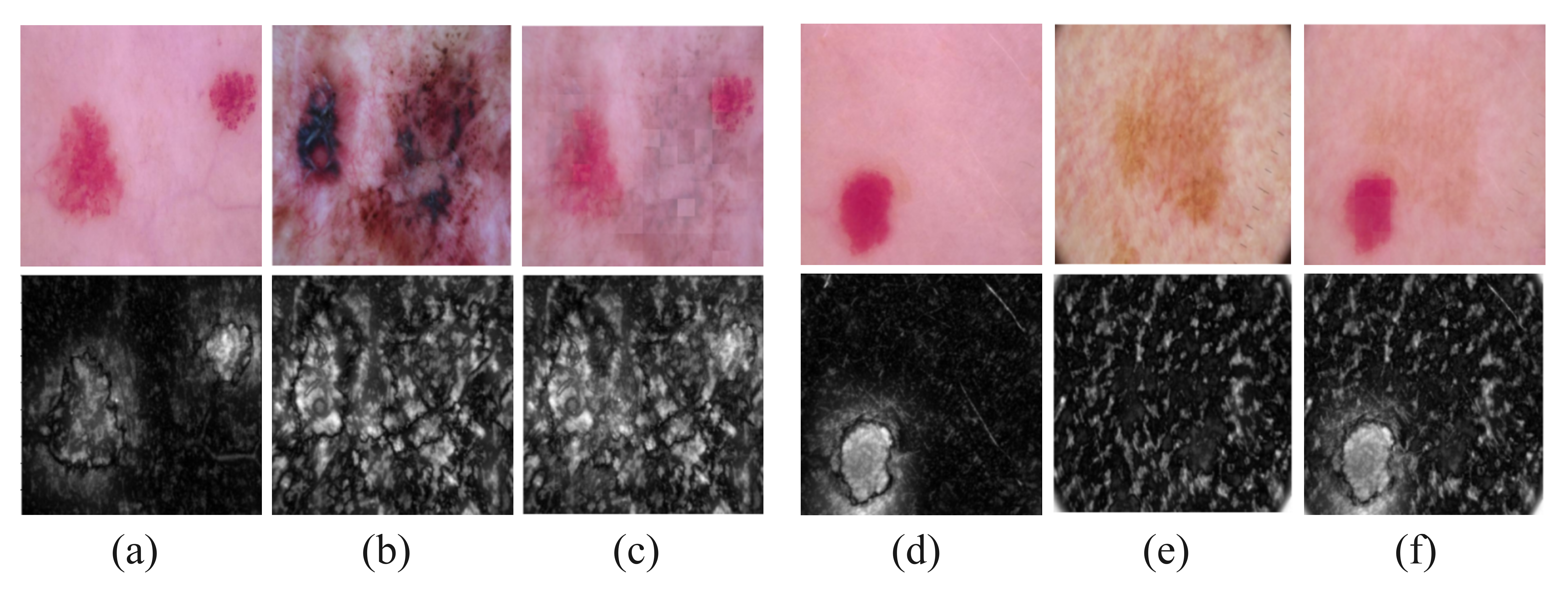}
\vspace{-0.9cm}
\caption{Visualization of augmented images generated by SPMix on the image level. The column (a) and (d) are the tail-class images. The column (b) and (e) are the head-class images. The column (c) and (f) are the augmented images.} \label{mixedsamples}
\vspace{-0.5cm}
\end{figure}

We employ the ISIC2018 dataset\cite{codella2019skin}, a largescale dataset of dermoscopy images. Following the experimental setup for long-tailed problems\cite{wang2021contrastive}, we split the original dataset into train, validation and test sets with a ratio of 7:1:2 and maintain balanced sample sizes for all classes in the validation and test sets for objective unbiased evaluation. Based on the number of samples in each category in the overall dataset the ranking, we categorize the ISIC2018 dataset into three subsets: “Many”, “Medium”, and “Few” for a more granular assessment (Fig.~\ref{dataset}). Accuracy is adopted to evaluate the performance of each subsets and F1 score is employed to evaluate the overall classification performance. We visualize the augmented images for the tail-classes Fig.~\ref{mixedsamples}. Our approach effectively replaces the background of the tail-class samples with the background of the head-class samples while preserving their distinctive features.

\begin{table}[t]
\centering
\caption{ Comparisons with state-of-the-art methods. Top-1 validation accuracy of ISIC2018 dataset. The optimal accuracy across all comparisons is denoted in \textbf{bold} while the second best ones are {\ul underlined}.}\label{tab1}
\vspace{0.1cm}
\resizebox{\linewidth}{!}
{

\begin{tabular}{@{}cccccc@{}}
\specialrule{0.10em}{0pt}{0pt}
\multicolumn{1}{l|}{\multirow{2}{*}{Method}} & \multicolumn{4}{c|}{Acc}                                  & \multirow{2}{*}{F1-score} \\
\multicolumn{1}{c|}{}                               & Many  & Med   & Few  & \multicolumn{1}{c|}{Total}  &                           \\ 
% \midrule
\specialrule{0.05em}{0pt}{3pt}
\multicolumn{6}{c}{\textbf{ResNet50}}                      \\ 
\specialrule{0.05em}{0pt}{3pt}
% \midrule
CE (baseline)                                        & 80.00 & 42.50 & 33.33 & 43.93                     & 43.13                     \\
CE-resample\cite{yang2022proco}                                            & 62.50 & 57.50 & 45.00  & 52.86                  & 52.51                     \\
MiSLAS \cite{zhong2021improving}                                            & 65.00 & 71.88 & 67.50    & 63.93                    & 64.87                     \\
ResLT \cite{cui2022reslt}                                                & 87.50 & 47.50 & 45.83      & 52.50               & 47.32                     \\
BALLAD  \cite{ma2021simple}                                    &   75.00            &  63.33    &   61.67              &      64.29         &       63.66             \\
BCL    \cite{zhu2022balanced}                                      & 80.00 & 79.17 & 65.83              & 73.57               &  74.11                         \\
PaCo      \cite{cui2021parametric}                                          & 15.00 & 59.17 & 81.67               & 62.50        & 60.21                     \\
% \midrule
\specialrule{0.05em}{0pt}{3pt}
\multicolumn{6}{c}{\textbf{Hybrid-ViT}}                      \\ 
\specialrule{0.05em}{0pt}{3pt}                                                 
% \midrule
CE (baseline)                                     & {\ul 90.00} & {\ul 90.00} & 58.33                  & 76.43       & 75.83                     \\
CE-resample\cite{yang2022proco}                            & \textbf{92.50} & 86.67 & 57.50                 & 75.00          & 74.25                     \\
MiSLAS     \cite{zhong2021improving}                                            & {\ul 90.00} & \textbf{90.62} & 75.00               & 75.00       & 74.12                     \\
BCL         \cite{zhu2022balanced}                                              &   {\ul 90.00}     &  85.83    &              63.33          &  76.79       &     76.55                      \\
PaCo            \cite{cui2021parametric}                                  & 72.50 & 86.67 & \textbf{86.67}            & {\ul 84.64}         & {\ul 84.59}               \\
Ours                                         &  82.50 & {\ul 90.00} & {\ul 83.33}                 & \textbf{86.07}    & \textbf{86.22}            \\ \bottomrule
\end{tabular}
}
\vspace{-0.5cm}
\end{table}
\vspace{-0.2cm}
\subsection{Implementation Details}
All our experiments are conducted on a workstation equipped with two NVIDIA RTX 3090 GPUs. Hybrid-ViT, which combines ResNet-50 with ViT-S, serves as the backbone. In this architecture, ResNet-50 extracts features, and ViT-S processes the mixed features as input, then producing the features for classification. The batch size is set to 64 with the initial learning rate of 5e-6. All models are trained using AdamW optimizer with momentum $\mu$ = 0.9 and the weight decay is 0.1. The temperature is set to 0.2. The input image size is set to be 224 × 224, and the feature dimension is 768. The best performance is achieved when the hyperparameter $\alpha$ = 0.8. The training epochs are set to be 500.
\vspace{-0.2cm}
\subsection{Comparison with State-of-the-art}
We compare our proposed method with state-of-the-art approaches on the ISIC2018 dataset. The experimental results, presented in Table~\ref{tab1}, showcase the performance of each method in terms of accuracy and F1-score. While Paco has improved the accuracy of the tail classes, it has significantly deteriorated the performance of the head classes. MiSLAS and BCL mainly focus on the performance of head classes. Our approach not only ensures the performance of the head classes but also enhances the accuracy of the tail classes. Notably, our method achieved significant accuracy improvements compared to the second-best approach. Specifically, we observed a substantial 1.43$\%$ accuracy and 1.63$\%$ F1-score improvement on the ISIC2018 dataset. 
\begin{table}[t]
\centering
\caption{Effectiveness of Patch-based Mixup and Saliency Guidance in our SPMix framework on ISIC2018. The best ones are \textbf{bolded} while the second best ones are {\ul underlined.}}\label{tab2}
\vspace{0.1cm}
\resizebox{\linewidth}{!}{
\begin{tabular}{@{}ccccc@{}}
\toprule
Patch-based mixup & Saliency guidance & Acc             & F1              \\ \midrule
 \XSolidBrush                 &         \XSolidBrush             & 83.57          & 83.47          \\
\Checkmark                 &        \XSolidBrush             & 83.93          & 84.08          \\
    \XSolidBrush                & \Checkmark                  & {\ul 84.64}          & {\ul 85.00}          \\
\Checkmark                  & \Checkmark                  & \textbf{86.07} & \textbf{86.22} \\ \bottomrule
\end{tabular}
}
\vspace{-0.5cm}
\end{table}
\vspace{-0.2cm}
\subsection{Ablation Study}
To evaluate the effectiveness of each component, we conduct ablation studies on the ISIC2018 dataset. As depicted in Table~\ref{tab2}, the introduction of patch-based mixup to mix the head-class samples and tail-class samples, the results have been improved to some extent. When traditional mixup is employed in conjunction with saliency guidance, a discernible benefit is observed. Finally, by integrating all three modules, we achieved the most substantial performance gains, resulting in performance enhancements of 2.5\% and 2.75\% respectively. 

\vspace{-0.2cm}

\section{CONCLUSION}

In this work, we propose a new framework, SPMix, to solve the long-tailed skin cancer classification problem. Our main idea is to compensate for the tail classes with the background of the head classes, and retain the key lesion information of the tail classes guided by the saliency map. Experiments are conducted on the ISIC2018 dataset, and the results fully demonstrate that SPMix achieves the best performance on the dataset compared with other long-tailed methods.

\small
% \vspace{-0.5cm}
\section{Acknowledgements}
% \vspace{-0.3cm}
This study was supported by the Shenzhen Basic Research Program (JCYJ20190809120205578); the National Natural Science Foundation of China (62071210); the Shenzhen Science and Technology Program (RCYX20210609103056042); the Shenzhen Basic Research Program (JCYJ20200925153847004); the Shenzhen Science and Technology Innovation Committee (KCXFZ2020122117340001).
% \vspace{-0.3cm}
%

% \vspace{-0.5cm}
\bibliographystyle{IEEEbib}
% \vspace{-0.5cm}
\bibliography{refs}
\end{document}